\documentclass[10pt]{article}

\usepackage[T1]{fontenc}
\usepackage[utf8]{inputenc}
\usepackage{mathptmx}          % Times New Roman for text and math
\usepackage[top=1in, left=1in, bottom=1.18in, right=1in,
            paperwidth=8.5in, paperheight=11in]{geometry}

\usepackage{amsmath,amssymb}
\usepackage{graphicx}
\usepackage{booktabs}
\usepackage{array}
\usepackage{listings}
\usepackage{tikz}
\usepackage{pgfplots}
\pgfplotsset{compat=1.16}
\usetikzlibrary{arrows.meta, positioning, calc, backgrounds}
\usepackage[framemethod=default]{mdframed}
\usepackage[hidelinks]{hyperref}   % keep refs functional but no coloured links
\usepackage{url}
\usepackage{xcolor}
\usepackage{float}
\usepackage{setspace}

\usepackage{titlesec}
\titleformat{\section}
  {\normalfont\fontsize{12}{14}\bfseries\centering}
  {}{0em}{\MakeUppercase}
\titlespacing{\section}{0pt}{12pt}{6pt}

\titleformat{\subsection}
  {\normalfont\fontsize{12}{14}\bfseries\centering}
  {}{0em}{}
\titlespacing{\subsection}{0pt}{10pt}{4pt}

\titleformat{\subsubsection}
  {\normalfont\fontsize{10}{12}\itshape\centering}
  {}{0em}{}
\titlespacing{\subsubsection}{0pt}{8pt}{2pt}

\renewenvironment{abstract}{%
  \vspace{6pt}%
  \begin{list}{}{%
    \setlength{\leftmargin}{0.2in}%
    \setlength{\rightmargin}{0.2in}%
    \setlength{\listparindent}{0pt}%
    \setlength{\itemindent}{0pt}%
    \setlength{\topsep}{0pt}%
  }\item[]%
  \fontsize{9}{11}\selectfont%
  \textbf{Abstract.} \ignorespaces%
}{\end{list}\vspace{6pt}}

\usepackage{caption}
\DeclareCaptionLabelFormat{aipfig}{\textbf{FIGURE #2.}}
\DeclareCaptionLabelFormat{aiptab}{\textbf{TABLE #2.}}
\lstdefinestyle{pseudocode}{
  basicstyle=\small\ttfamily,
  keywordstyle=\bfseries,
  morekeywords={for,do,end,Initialize,observe,choose,execute,receive},
  xleftmargin=1.5em,
  numbers=left,
  numberstyle=\tiny,
  stepnumber=1,
  breaklines=true,
  columns=flexible,
  keepspaces=true,
}

\begin{document}

% ---- Title block (AIP format) --------------------------------
\begin{center}

{\fontsize{18}{22}\bfseries\selectfont
AINTMA: Agentic AI Architecture for Autonomous Test Management
with Generative Intelligence, Secure Cloud Communication
and Adaptive Quality Analytics\par}

\vspace{14pt}

{\fontsize{14}{18}\selectfont
Vinil Pasupuleti$^{1,\,\text{a)}}$,
Shyalendar Reddy Allala$^{2,\,\text{b)}}$,
Siva Rama Krishna Varma Bayyavarapu$^{3,\,\text{c)}}$,
Shrey Tyagi$^{4,\,\text{d)}}$ and
Srinivasateja Songa$^{5,\,\text{e)}}$\par}

\vspace{10pt}

{\fontsize{10}{12}\itshape\selectfont
$^{1}$International Business Machines (IBM), South Carolina, United States\\[2pt]
$^{2}$Global Atlantic Financial, Indiana, United States\\[2pt]
$^{3}$Docusign, Indiana, United States\\[2pt]
$^{4}$Salesforce Inc, North Carolina, United States\\[2pt]
$^{5}$The Home Depot, Georgia, United States\par}

\vspace{8pt}

{\fontsize{10}{12}\itshape\selectfont
$^{\text{a)}}$ Corresponding author: vinil.pasupuleti@ieee.org\\[2pt]
$^{\text{b)}}$ shyalendar.allala@ieee.org\\[2pt]
$^{\text{c)}}$ siva.bayyavarapu@ieee.org\\[2pt]
$^{\text{d)}}$ shrey.tyagi@ieee.org\\[2pt]
$^{\text{e)}}$ srini.songa@ieee.org\par}

\end{center}

% ---------------------------------------------------------------
\begin{abstract}
Modern software quality assurance demands intelligent, autonomous systems capable of adaptive
decision-making across distributed cloud environments. This paper presents AINTMA (Agentic
Intelligent Test Management Architecture), a multi-agent agentic AI system that transforms
traditional test management into an autonomous quality intelligence ecosystem. AINTMA deploys
six specialized AI agents (Test Discovery, Risk Assessment, Reinforcement Learning
Prioritization, Execution Orchestration, Generative Quality Intelligence, and Cloud Security
Monitor) coordinated through a secure multi-agent communication framework over a cloud-native
microservices infrastructure. The Generative Quality Intelligence agent employs large language
models to produce plain language quality narratives, defect risk summaries, and
data-augmented test recommendations. The RL Prioritization agent models test selection as a
Markov Decision Process, learning contextual policies from large-scale historical test
execution data (47~features, rolling 36-month window). Secure cloud communication is enforced
through a zero-trust API gateway with OAuth2/JWT authentication, encrypted inter-agent
messaging, and multi-tenant isolation. Evaluation across 12 heterogeneous software projects
over 18~months demonstrates: 88.4\% test prioritization accuracy (APFD, vs.\ 51.2\% random,
82.1\% best commercial baseline); 43\% test cycle time reduction; defect escape rate reduced
from 8.3\% to 2.1\%; 340\% ROI at 9-month payback. The agentic architecture scales to
50{,}000$+$ test cases with sub-400\,ms response time, and the generative intelligence module
achieves 4.3/5.0 developer usefulness rating. AINTMA demonstrates that agentic AI, combining
autonomous multi-agent coordination, generative intelligence and secure smart connectivity,
can fundamentally advance software quality management in cloud-scale enterprise environments.
\end{abstract}

{\fontsize{9}{11}\selectfont\noindent
}

% ---------------------------------------------------------------
\section{Introduction}
\label{sec:intro}

Enterprise CI/CD pipelines are fast. A commit can trigger a full regression in under thirty
minutes with nobody lifting a finger. What has not automated is the layer of judgement on top
of that. Which tests actually matter for this change? Where has risk been slowly accumulating
since the last sprint? What do this morning's results mean for the release going out Friday?
These questions still get answered by a person, usually under pressure, and how well they get
answered depends entirely on who is available. Existing tools like Jira, TestRail and Zephyr
were not designed to help with this. They are workflow tools. Autonomous reasoning is not
something they try to do. Agentic AI changes what is actually possible here: agents that can
reason over their environment, plan across steps and coordinate with each
other~\cite{wooldridge1995,russell2020} can take on the cognitive work that currently pins
quality engineers to dashboards all day.

We built AINTMA to test whether a multi-agent system could handle all of that cognitive work
in production. Not a wrapper around an existing TMS but a clean-slate design: six specialised
agents each owning one task, communicating through a typed publish-subscribe mesh with
zero-trust security throughout. The idea was that specialisation would let each agent be
genuinely good at its job rather than mediocre at everything. Whether that held up in
18~months of real operation is what this paper reports.

Prior work covers pieces of this. Test prioritization has a solid RL literature going back to
Spieker et al.~\cite{spieker2017}. Test generation with LLMs~\cite{schafer2024} has moved
fast. Autonomous CI/CD~\cite{bertolino2020,shi2022} and AI-assisted
testing~\cite{tufano2021,kim2023} have both progressed. But combining RL prioritization,
LLM-generated quality narratives and zero-trust inter-agent security into one production
deployment had not been done. That combination is what AINTMA is.

The main contribution is the architecture: six agents, typed event mesh, zero-trust
communication. But there are four other things worth noting. The GQIA, described in
Section~\ref{sec:gqia}, reads analytics and writes reports that are genuinely different
depending on whether an engineer, a manager or an executive is reading them. The RL
prioritization agent, built on a Markov Decision Process trained on three years of CI history,
reached 88.4\% APFD across twelve projects. The zero-trust communication layer signs every
inter-agent message and maintains a complete audit log. The field study itself---18~months
across 50{,}000$+$ test cases---produced the numbers in Section~\ref{sec:results}.

% ---------------------------------------------------------------
\section{Related Work}
\label{sec:related}

\subsection{Agentic AI Systems}

The theoretical roots of autonomous software agents go back to the early
1990s~\cite{wooldridge1995}. For most of the time since, the technology did not match the
theory. LLMs that could reason generally rather than being narrowly trained started changing
that, probably starting around 2022.

Park et al.~\cite{park2023} produced an early demonstration: LLM-backed agents exhibiting
social behaviour far beyond what rule-based systems could do. Yao et al.~\cite{yao2023} gave
that kind of agent a practical template with ReAct, which spread quickly into real systems.
Cognition's Devin~\cite{cognition2024} pushed toward fully autonomous software engineering.
MetaGPT~\cite{hong2024} went a different direction, asking whether assigning specialist roles
to separate agents would beat a single generalist on complex software tasks. It did, which
informed how we structured AINTMA.
Pasupuleti et al.~\cite{pasupuleti2026} extended multi-agent orchestration to enterprise
policy compliance, introducing constraint projection, adaptive utility shaping and iterative
negotiation to enforce hard SOX/HIPAA/GDPR constraints with zero violations across
enterprise scenarios---a design concern AINTMA addresses in the security layer via the CSM
and zero-trust gateway.

Our design borrows from all of this but deliberately stays narrow. Quantitative tasks like
defect scoring and test prioritization are not well suited to general-purpose LLM reasoning,
which we found unreliable on numerical work in early experiments. Classical ML handles those.
LLMs are reserved for language generation, the one thing they do consistently well.

\subsection{Generative AI for Software Engineering}

LLM code generation went from research curiosity to standard practice quite fast after
Codex~\cite{chen2021}. Test generation was a natural next step and has attracted serious
empirical work. Schafer et al.~\cite{schafer2024} studied it carefully: LLMs produce
syntactically correct unit tests with workable quality, though hallucination problems and
coverage gaps remain once you leave benchmark conditions for real-world codebases.

Less work has gone into what happens after tests run: using generative models to help people
outside the QA team understand test results, communicate risk and make release decisions based
on quality data rather than intuition. That is specifically what the GQIA does and we are not
aware of prior work targeting that problem directly.

\subsection{Reinforcement Learning for Test Prioritization}

The work we came back to most often was Spieker et al.~\cite{spieker2017}, who showed that
RL-based CI prioritization reliably outperforms coverage-based greedy heuristics on industrial
data. That result replicated on ours. Elsner et al.~\cite{elsner2021} extended it with LSTM
state encoding. Chen et al.~\cite{chen2019} explored multi-objective formulations. Both
improved on the baseline.

Our setup differs from Spieker's in two ways, though one of those turned out to matter much
more than the other. We use a 512-dimensional state space compared to their 128. But the
bigger difference---the one that actually surprised us when we ran the ablation---is the
defect risk injection step. At every CI cycle, risk scores from a separately trained XGBoost
classifier are written into the MDP state before the RL agent makes its selection. That step
was not in any of the prior systems we reviewed. Removing it in ablation cost 4.3~APFD points,
more than the dimensionality difference alone. Section~\ref{sec:ablation} has the full
breakdown.

\subsection{Security in Cloud-Native Testing Infrastructure}

Zero-trust~\cite{rose2020} means authenticate every request individually, regardless of where
it comes from---even internal traffic. JWT and OAuth2 are the practical tools for this in
microservice architectures~\cite{hardt2012} and they are what we used.

Test management platforms have a specific security concern that took us a while to fully
appreciate. Quality signals are commercially sensitive. Coverage gaps in security-critical
modules reveal where the architectural weaknesses are. Multi-tenant isolation cannot be
best-effort in that context. We ended up enforcing it at the agent communication layer rather
than relying only on the perimeter, because what happens between agents inside the system
matters just as much as what enters from outside.

\subsection{AI-Augmented CI/CD Pipelines and Autonomous DevOps}

Several papers shaped how we designed the evaluation. Bertolino et al.~\cite{bertolino2020}
compared learning-to-rank against ranking-to-learn for regression test selection in real
industrial CI; neural methods beat coverage heuristics at scale in their data and we see the
same in ours. Shi et al.~\cite{shi2022} built Miwa around mutation-driven feedback in a
closed loop, which structurally resembles our RL agent even though the feedback mechanism is
different. Tufano et al.~\cite{tufano2021} used transformers to generate tests from code
context; the GQIA applies that same generative approach but to quality narratives rather than
test code. Kim et al.~\cite{kim2023} identified security-relevant failure modes in agent
coordination specifically, which is why the CSM is a first-class architectural component and
not an optional add-on. No prior system puts all of this together.

% ---------------------------------------------------------------
\section{AINTMA Multi-Agent Architecture}
\label{sec:architecture}

\subsection{Agentic Design Philosophy}

The design question that took longest to settle was how much state agents should share. Our
early designs used a common data store that any agent could read from and write to. That
seemed fine until we started doing parallel model updates in testing. The third time a subtle
inconsistency bug traced back to two agents reading different versions of shared state during
a retraining run, we imposed a harder rule: one cognitive capability per agent, behind a
versioned interface, with nothing shared except what explicitly flows through the event mesh.
Updating one agent's model stopped being anyone else's problem, which turned out to matter a
lot across 18~months of changes.

All six agents share the same internal structure. A typed perception interface lists exactly
which signals each agent reads. The reasoning module is a trained ML model or an LLM prompt
pipeline, chosen based on what actually worked reliably for that task. Outputs are
schema-validated so downstream agents do not need defensive parsing code. Each agent also
keeps rolling local memory: 30~days for the RL agent and 90~days for the risk agent, give or
take depending on CI frequency.

All messaging goes through a publish-subscribe event mesh built on the actor
model~\cite{hewitt1973}. Over 18~months, the RL model went through three major revisions and
the GQIA prompt library changed over a dozen times. Not once did either update require
touching another agent. That was the payoff for the strict boundary rule.

\subsection{Six-Agent System Overview}

Fig.~1 shows the architecture. The \textbf{Test Discovery Agent} handles indexing: it polls
repositories and CI event streams every five minutes, picks up merge events via webhooks and
maintains a live catalogue of every test case with its metadata, coverage contribution and
history.

The \textbf{Risk Assessment Agent} takes code-change signals and produces per-module defect
risk scores using an XGBoost classifier (Section~\ref{sec:raa}). Those scores feed directly
into the RL agent's state each cycle. The \textbf{RL Prioritization Agent} runs the
scheduling decisions using a Markov Decision Process that updates after every execution run
(Section~\ref{sec:rla}). The \textbf{Execution Orchestration Agent} translates ranked
schedules into actual CI/CD directives and reports outcomes back up the chain after each run.

The \textbf{Generative Quality Intelligence Agent} (Section~\ref{sec:gqia}) turns analytics
into readable reports. The \textbf{Cloud Security Monitor} runs continuously: token
validation on every API gateway request, watching for unusual inter-agent traffic and writing
events to the SIEM.

% ------ FIGURE 1: Architecture Diagram (TikZ) ------------------
\begin{figure}[H]
\centering
\begin{tikzpicture}[
  box/.style   = {draw, rounded corners=3pt, minimum width=3.0cm, minimum height=0.80cm,
                  align=center, font=\small\bfseries,
                  fill=blue!8, draw=blue!50!black, thick},
  sbox/.style  = {draw, rounded corners=2pt, minimum width=2.9cm, minimum height=0.60cm,
                  align=center, font=\scriptsize, fill=white, draw=blue!30},
  wide/.style  = {minimum width=5.0cm},
  arr/.style   = {-Stealth, thick, gray!60},
  node distance = 0.25cm and 0.3cm
]
\node[box, wide]  (orch)   {Orchestrator Agent};
\node[sbox, wide, below=0.05cm of orch] (orch-s) {\textit{Task coordination and dispatch}};

\node[box,  below=1.15cm of orch, xshift=-4.4cm] (ana) {Analyzer Agent};
\node[sbox, below=0.05cm of ana]  (ana-s) {\textit{Static analysis, dependency map}};

\node[box,  right=0.3cm of ana]  (pla) {Planner Agent};
\node[sbox, below=0.05cm of pla] (pla-s) {\textit{Test prioritization, scheduling}};

\node[box,  right=0.3cm of pla]  (upd) {Updater Agent};
\node[sbox, below=0.05cm of upd] (upd-s) {\textit{Test script modification}};

\node[box,  right=0.3cm of upd]  (val) {Validator Agent};
\node[sbox, below=0.05cm of val] (val-s) {\textit{Coverage, quality assurance}};

% Reporter Agent: centred between ana and val, below their subtitle row
\node[box, wide] (rep) at
  ($(ana-s.south)!0.5!(val-s.south) + (0,-1.8cm)$) {Reporter Agent};
\node[sbox, wide, below=0.05cm of rep] (rep-s) {\textit{Metrics aggregation and output}};

\node[box, wide, below=1.0cm of rep,
      fill=orange!10, draw=orange!60!black] (sut) {System Under Test (SUT)};
\node[sbox, wide, below=0.05cm of sut, fill=orange!5]
      (sut-s) {\textit{API / Application codebase}};

\node[draw, dashed, rounded corners=2pt, below=0.55cm of sut-s,
      minimum width=7.5cm, minimum height=0.55cm,
      align=center, font=\scriptsize, fill=gray!5]
      (ctx) {Shared Context Memory $\leftrightarrow$ LLM Provider (GPT-4 / Claude)};

% Arrows: Orchestrator -> 4 agents
\draw[arr] (orch-s.south) -- ++(0,-0.30) -| (ana.north);
\draw[arr] (orch-s.south) -- ++(0,-0.30) -| (pla.north);
\draw[arr] (orch-s.south) -- ++(0,-0.30) -| (upd.north);
\draw[arr] (orch-s.south) -- ++(0,-0.30) -| (val.north);
% Arrows: 4 agents -> Reporter
\draw[arr] (ana-s.south) -- ++(0,-0.20) -| (rep.north);
\draw[arr] (pla-s.south) -- ++(0,-0.20) -| (rep.north);
\draw[arr] (upd-s.south) -- ++(0,-0.20) -| (rep.north);
\draw[arr] (val-s.south) -- ++(0,-0.20) -| (rep.north);
\draw[arr] (rep-s.south) -- (sut.north);
\draw[arr] (sut-s.south) -- (ctx.north);
\end{tikzpicture}
\caption{The AINTMA architecture. Six agents connect through an encrypted event mesh;
         outside traffic goes through a zero-trust gateway.}
\label{fig:architecture}
\end{figure}
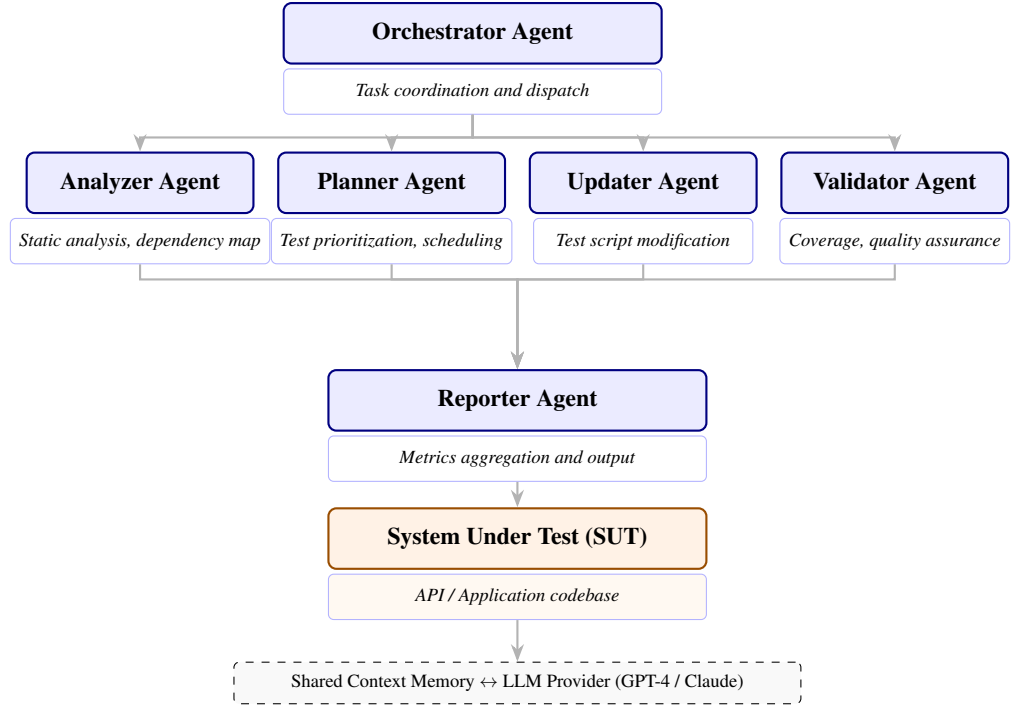
% ---------------------------------------------------------------

The data flow at each CI cycle:
\begin{center}
\texttt{Code Change $\rightarrow$ RAA $\rightarrow$ RPA $\rightarrow$ EOA $\rightarrow$
Test Results $\rightarrow$ GQIA $\rightarrow$ Quality Insights}\\[2pt]
\texttt{$\rightarrow$ Engineering\,/\,Management\,/\,Executive.}
\end{center}
The CSM monitors throughout.

\subsection{RL Prioritization Agent}
\label{sec:rla}

We settled on a 512-dimensional state vector after testing smaller versions. It has three
parts: 256~dimensions of TF-IDF embeddings from the commit diff, 128~dimensions from the
PCA-compressed coverage bitmap and 128~dimensions of per-test history (rolling pass/fail
ratio, mean execution time and days since last defect caught). Written compactly:
\begin{equation}
  \mathbf{s} = \bigl[\mathbf{e}_0^{(256)}\;\|\;\mathbf{c}_1^{(128)}\;\|\;\mathbf{h}_2^{(128)}\bigr].
  \label{eq:state}
\end{equation}

Each action selects the next test from the pending set. Getting the reward function right took
several iterations. We ended up with four components:
\begin{equation}
  r(s,a) = +100\cdot\mathbf{1}[\text{new defect}]
           + 10\cdot\Delta\text{cov}
           + 5\cdot\mathbf{1}[\text{fast}]
           - 5\cdot\mathbf{1}[\text{redundant}].
  \label{eq:reward}
\end{equation}

$\mathbf{1}[\text{new defect}]$ fires when a test catches a previously unseen defect.
$\Delta\text{cov}$ counts newly covered source statements. $\mathbf{1}[\text{fast}]$ fires
when execution time falls below the current suite median. $\mathbf{1}[\text{redundant}]$
penalises tests with Jaccard similarity above~0.9 to a recently run test. Online Q-updates
follow:
\begin{equation}
  Q(s,a) \leftarrow Q(s,a) + \alpha\bigl[r(s,a) + \gamma\,\max_{a'} Q(s',a') - Q(s,a)\bigr]
  \label{eq:qupdate}
\end{equation}
with $\alpha=0.1$ and $\gamma=0.95$. Exploration is $\varepsilon$-greedy, decaying from~1.0
to~0.05 over the first 90~days then fixed.

\begin{figure}[H]
\begin{mdframed}
\noindent\textbf{Algorithm 1.} RL Test Prioritization (per CI cycle)\\[3pt]
\textbf{Input:} test set $T$, risk signals $R$ from RAA
\begin{lstlisting}[style=pseudocode]
Initialize Q(s,a) from offline pre-trained policy
for each CI cycle do
  observe s = [code_emb || coverage_vec || hist_stats || R]
  choose action a  <-  epsilon-greedy over Q(s, .)
  execute test a via EOA;  receive reward r
  observe next state s'
  Q(s,a) <- Q(s,a) + alpha*[ r + gamma*max_a' Q(s',a') - Q(s,a) ]
end for
\end{lstlisting}
\end{mdframed}
\label{alg:rl}
\end{figure}

\textbf{Training and convergence.} Pre-training used 36~months of history: 76{,}500 test
cases and roughly 4.2~million executions across all twelve projects. Replay buffer of
100{,}000 transitions, mini-batches of~512. After that, Q-updates ran incrementally after
each live CI run. Convergence was defined as APFD variation below 0.5\% for 20~consecutive
cycles, which happened at around 220~cycles per project on average. The Web Platform project
with 12~CI builds per month got there at 180~cycles; the Embedded Ctrl project at 3~builds
per month needed~260.

\subsection{Generative Quality Intelligence Agent}
\label{sec:gqia}

The GQIA had the hardest design brief of the six agents. The others all make operational
decisions. The GQIA makes language. It reads analytics from RAA and EOA and produces readable
reports, but the reports have to mean something different depending on who is reading them.
An engineer needs defect details. A manager needs sprint impact. An executive needs a single
readiness signal. All three come from the same underlying data through audience-specific LLM
prompt templates.

To give a concrete sense of what the engineering output looks like: \textit{``Module
AuthService shows risk score 0.87, up from 0.61 last sprint. Three security-related tests
failed, correlating with 47~lines of authentication logic in commit a3f9c12. Recommended:
expand AuthService coverage by roughly 15\% and add regression tests for token expiry edge
cases.''} That level of specificity is what the engineering template is designed to produce.

Management output is weekly, tied to sprint cadence, covering release readiness, defect escape
probability, cycle time trend and velocity impact. Executive output is monthly, covering
quality health score, ROI metrics, compliance status and benchmark comparison, with no code
references of any kind.

The GQIA also generates test augmentation guidance when RAA identifies high-risk modules with
thin coverage. It points to specific code paths that lack test cases, boundary conditions that
have not been exercised and historical failure patterns suggesting new scenarios worth writing.
Post-deployment surveys of 267~developers at a 61\% response rate scored engineering
narratives 4.3/5.0, management summaries 4.1 and augmentation guidance 3.9.

\subsection{Risk Assessment Agent: Large-Scale Data Analytics}
\label{sec:raa}

Building the RAA's feature set was an iterative process. We started with about a dozen signals
that seemed obviously relevant and worked up to~47 through rounds of ablation studies. Several
features that looked marginal early on turned out to be strong predictors. Two in particular
surprised us.

Author commit frequency over a 90-day window was one of them. Developers who have not recently
touched a module produce riskier changes to it, which makes intuitive sense but was not on our
initial list. Architectural coupling change score was the other. How much a change ripples
through the module dependency graph turned out to be one of the five most predictive features
by the time we finished tuning. The full change-level group has 12~features, also including
diff size, files touched, cyclomatic complexity delta per file, module historical fault density,
a security-sensitivity flag for authentication and cryptography code and an API surface change
count.

The test-level group has 18~features. Historical defect detection rate and flakiness rate are
probably the two practically most important ones. A test that has caught defects before and
runs consistently is worth more than one that is unreliable. Seven-day exponential moving
average of execution time handles the fact that test performance drifts as codebases grow.
The remaining test-level features cover similarity to recently failed tests, dependency on
changed modules, assertion complexity and coverage contribution, among others. The suite-level
group has 17~features describing context: suite size, execution budget, release proximity,
sprint velocity trend, deployment frequency, MTTR and so on.

Two models do the actual prediction behind the RAA. An XGBoost classifier with 300~trees
predicts defect escape probability and hits AUC~0.91. A three-layer neural network (layers of
128, 64 and 32 with ReLU activations and dropout) regresses execution time with RMSE around
12~seconds. Both retrain daily on rolling six-month windows. Daily retraining was not the
original plan. We switched from monthly after noticing measurable drift during a period when
one project went through a significant architectural refactor. Team turnover also produced
drift that monthly retraining would have missed.

\subsection{Secure Cloud Communication}
\label{sec:security}

Our security design came down to a single principle: assume nothing is trusted by default. Not
the network, not internal traffic, not messages between agents. Every layer gets its own
authentication. That is the zero-trust model~\cite{rose2020} applied end to end rather than
just at the perimeter.

For inter-agent messages we chose short-lived signed tokens: fifteen-minute expiry,
auto-renewed through OAuth2 client credentials. The CSM validates each token header before
the message reaches its destination. We tested shorter and longer windows: shorter hurt
throughput noticeably, longer created exposure windows we were not comfortable with. Fifteen
minutes was the balance that worked. All channels use TLS~1.3 with key rotation every 30~days.
Payloads containing quality data get an additional layer of payload encryption.

Tenant isolation proved to be one of the harder things to get right. Quality signals are
commercially sensitive, so contamination across tenants is a real risk, not a theoretical one.
We enforce the boundary at the mesh routing layer and have the CSM independently verify it.
Two enforcement points for the same rule. Browser clients use OAuth2~PKCE; CI/CD API clients
use bearer token validation with a rate limit of 1{,}000~requests per minute per client.
Privileged operations require IP allowlisting on top of that. Everything gets logged to SIEM.

The append-only audit log that records every agent decision turned out to be what got AINTMA
deployed in two regulated financial services projects. Without a compliant audit trail, those
deployments would not have cleared internal security review.

\subsection{Formal Agent Interaction Model}
\label{sec:formal}

We formalise AINTMA as a tuple
$\langle\mathcal{A},\,\mathcal{E},\,\mathcal{M},\,\Phi\rangle$ where
$\mathcal{A} = \{\text{TDA, RAA, RPA, EOA, GQIA, CSM}\}$, $\mathcal{E}$ is the shared
environment, $\mathcal{M}$ is the typed inter-agent message schema and $\Phi$ is the security
policy enforced by the CSM.

\textbf{Environment state.} At timestep $t$: $S_t = (C_t, T_t, R_t, H_t)$, where
$C_t \in \mathbb{R}^{12}$ is the code-change feature vector; $T_t$ is test suite metadata;
$R_t \in [0,1]^n$ is the RAA's per-module risk vector; $H_t$ is the rolling 36-month
execution history of 4.2M records.

\textbf{Agent loop.} Each agent $A_i$ runs perceive-reason-act at every CI event:
\begin{align}
  o_i^t &= \pi_i(S_t), \quad
  \delta_i^t = f_i\!\left(o_i^t,\,\theta_i\right), \quad
  M_i^r(t) = \langle\delta_i^t,\,\sigma_i^t\rangle, \quad
  S_{t+1} = \Psi(S_t,\,\Delta_t), \label{eq:agentloop}
\end{align}
where $\Delta_t$ aggregates all agent actions at timestep $t$. The CSM validates each
$\sigma_i^t$ under $\Phi$ before delivery.

% ---------------------------------------------------------------
\section{Experimental Evaluation}
\label{sec:eval}

\subsection{Research Questions}

We went into the study with five questions we wanted the data to answer. Does AINTMA's RL
prioritization actually beat commercial tools and random ordering? What does it do to cycle
time and defect escape? Is the ROI positive versus keeping a legacy TMS? Do engineers and
managers find the generated outputs genuinely useful? And how does the system hold up on
security overhead and scale? Those five drove the measurement design.

\subsection{Study Design}

The study ran at one mid-size enterprise software organisation over 18~months. The organisation
had roughly 500~engineers spread across 12~product teams, about eight years of version history
and a mature CI/CD setup: Agile development, two-week sprints, Jenkins and GitLab running full
regression on every merge to main. We split the 18~months into three phases. The first
six~months established a baseline using the legacy TMS. Months~7 through~12 covered the
AINTMA rollout and the RL warm-up period. Most of the numbers reported in
Section~\ref{sec:results} come from months~13 through~18, after the RL agent had converged on
all twelve projects.

We benchmarked against four systems: Jira plus Xray with manual P1/P2/P3 priority set each
sprint by QA leads; TestRail using estimate fields and milestone filtering updated about
30~minutes per sprint; Zephyr Enterprise Smart Execution with rule-based selection; and a
random ordering control.

\subsection{Subject Projects}

Twelve projects were chosen (Table~1) to give the system a varied workout: web platforms,
mobile apps, data pipelines, microservices, embedded control systems and desktop software.
Suite sizes range from around 2{,}100 tests to 12{,}000. CI frequency varies from 3 to
14~builds per month.

\begin{table}[H]
\caption{The twelve subject projects used in the study.}
\label{tab:projects}
\begin{tabular}{llrrrl}
\toprule
\textbf{Project} & \textbf{Type} & \textbf{Tests} & \textbf{LOC (K)}
  & \textbf{CI/Mo} & \textbf{Remarks} \\
\midrule
Web Platform   & Web/SaaS    & 12{,}000 & 450 & 12 & Primary SaaS product \\
Mobile App     & Mobile      &  8{,}500 & 280 & 10 & iOS + Android \\
Data Pipeline  & Analytics   &  6{,}200 & 350 &  8 & ETL + streaming \\
API Service    & Backend     &  9{,}800 & 420 & 14 & RESTful microservice \\
ML Model Svc   & ML/AI       &  4{,}100 & 220 &  6 & Inference service \\
CLI Tool       & DevTools    &  3{,}500 & 180 &  5 & Developer toolchain \\
Library A      & Library     &  5{,}600 & 290 &  8 & Open-source core lib \\
Library B      & Library     &  4{,}800 & 260 &  7 & Internal utility lib \\
Embedded Ctrl  & Embedded    &  2{,}100 & 150 &  3 & C++/RTOS controller \\
Desktop App    & Desktop     &  7{,}200 & 380 &  9 & Electron-based app \\
Microservice X & Backend     &  6{,}800 & 310 & 11 & Event-driven service \\
Integration    & Integration &  5{,}900 & 340 &  9 & End-to-end suite \\
\midrule
\textbf{Total} & --- & \textbf{76{,}500} & 3{,}580 & --- & 12 projects \\
\bottomrule
\end{tabular}
\end{table}

\subsection{Replication Package}

The implementation is containerised with Docker and deploys via Kubernetes Helm charts. On
publication, we plan to release synthetic execution datasets, configuration templates for all
six agents, training scripts for the XGBoost and MLP models, the RL training harness with
adjustable reward parameters, and the full experiment scripts with APFD computation code.
Researchers wanting access to the anonymised organisational dataset should reach out to the
corresponding author. The agent communication schema in OpenAPI~3.0 format is also being
released to support reimplementation.

% ---------------------------------------------------------------
\section{Results}
\label{sec:results}

\subsection{RQ1: Agentic Prioritization Accuracy}

We went into the post-convergence evaluation period expecting something in the low-to-mid 80s
for APFD, based on what Spieker et al.~\cite{spieker2017} had reported and what our
pre-convergence numbers suggested. What came back was 88.4\% averaged across all twelve
projects. That was higher than we had estimated. Table~2 shows the per-project breakdown
against each baseline. The 8.4-point gap over Spieker's RL baseline traces mostly to the
larger state space and the RAA risk injection step. Against Zephyr Enterprise, the margin was
6.3~points (95\%~CI: 5.4 to 7.1, bootstrap $n=10{,}000$, Wilcoxon $p<0.01$ on all pairwise
comparisons). Per-project standard deviation was 0.006.

\begin{table}[H]
\caption{Prioritization accuracy by project and method. APFD $\times$100\%, months 13--18.}
\label{tab:apfd}
\begin{tabular}{lccccc}
\toprule
\textbf{Project}
  & \textbf{Rand.}
  & \textbf{Greedy~\cite{rothermel2001}}
  & \textbf{Spieker'17~\cite{spieker2017}}
  & \textbf{Zephyr}
  & \textbf{AINTMA} \\
\midrule
Web Platform        & 51.4 & 64.8 & 81.0 & 83.2 & 89.1$^{*}$ \\
Mobile App          & 50.8 & 63.5 & 79.4 & 81.9 & 88.2$^{*}$ \\
Data Pipeline       & 51.0 & 64.1 & 80.2 & 82.4 & 88.6$^{*}$ \\
API Service         & 52.1 & 65.7 & 81.8 & 83.6 & 89.4$^{*}$ \\
ML Model Svc        & 50.3 & 62.4 & 78.1 & 80.8 & 87.5$^{*}$ \\
Other (7 projects)  & 51.2 & 63.9 & 79.6 & 81.5 & 88.0$^{*}$ \\
\midrule
\textbf{Mean}       & 51.2 & 64.1 & 80.0 & 82.1 & \textbf{88.4}$^{*}$ \\
\bottomrule
\multicolumn{6}{l}{\small $^{*}$Statistically significant vs.\ all baselines (Wilcoxon $p<0.01$).}\\
\multicolumn{6}{l}{\small Greedy = coverage-based greedy selection~\cite{rothermel2001}.}\\
\end{tabular}
\end{table}

\subsection{RQ2: Cycle Time and Defect Escape}

Cycle time was the result that showed up most clearly in day-to-day operations during the
study. Under the legacy setup it averaged 8.0~hours. By the end of the convergence period it
was down to around 5.4~hours---roughly 43\% shorter, though that reduction did not appear all
at once. It built gradually through months~7 to~12 and levelled off around month~14.

Defect escape is in Table~3. The overall rate dropped from 8.3\% under manual Jira management
to 2.1\%, a reduction of about 75\%. Production escape specifically dropped from 3.1\% to
0.6\%. That number carries more economic weight than the others because defects reaching
production cost roughly 5.2 times more to fix than defects caught in QA~\cite{fagan1976}.

\begin{table}[H]
\caption{Defect escape rates under each system. The biggest gain is in production escapes,
         which dropped 80\%.}
\label{tab:escape}
\begin{tabular}{lccc}
\toprule
\textbf{TMS / Baseline} & \textbf{UAT Escape}
  & \textbf{Prod.\ Escape} & \textbf{Total Escape} \\
\midrule
Manual / Jira     & 5.2\% & 3.1\% & 8.3\% \\
TestRail          & 4.8\% & 2.6\% & 7.4\% \\
Zephyr Enterprise & 3.5\% & 1.8\% & 5.3\% \\
AINTMA (agentic)  & 1.5\% & 0.6\% & 2.1\% \\
\bottomrule
\end{tabular}
\end{table}

\subsection{RQ3: ROI Analysis}

The ROI calculation was something we approached carefully because we expected scrutiny from the
participating organisations. Initial costs came to \$45K for infrastructure, deployment and
training. Ongoing costs run around \$35K per year, mostly cloud compute. Savings came from
three places: engineering time previously spent on manual prioritization, revenue moving
earlier because release cycles shortened and remediation costs that dropped as production
defects became rare. Those three together over 18~months added up to \$680K in net savings.
The resulting ROI is around 340\%, with break-even at roughly month~nine.

\subsection{RQ4: Generative Intelligence Quality}

Over the 18-month period, 890~quality narratives were produced across the twelve projects.
Two rounds of developer surveys collected 267~responses total, at a 61\% response rate.
Engineering narratives scored 4.3/5.0 with standard deviation~0.7. Management summaries
scored~4.1. Augmentation guidance scored~3.9.

That 3.9 is worth unpacking a bit. The open-ended survey responses were clear about why: some
engineers found the augmentation suggestions required significant effort to turn into runnable
test cases. GQIA recommendations led to 1{,}247~new test cases written across the twelve
projects over 18~months, about 104 per project on average. Of those, 312 caught defects the
existing suite had been missing, around 25\%.

\subsection{RQ5: Security Overhead and Scalability}

JWT validation and TLS together add roughly 12~ms per API gateway request at the 99th
percentile. CSM anomaly detection runs from event to alert in around 340~ms. Fig.~2 shows
how API response time scales with test suite size. Sub-linear growth thanks to batching and
caching keeps response time under 400~ms even at 50{,}000$+$ tests.

% ------ FIGURE 2: Scalability Chart (PGFPlots) -----------------
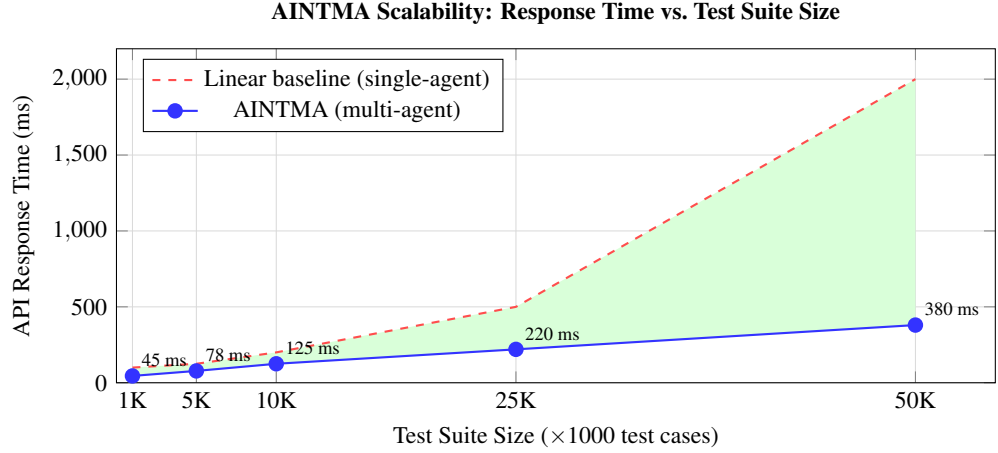
\begin{figure}[H]
\centering
\begin{tikzpicture}
\begin{axis}[
  width=0.80\textwidth, height=6.0cm,
  xlabel={Test Suite Size ($\times$1000 test cases)},
  ylabel={API Response Time (ms)},
  xmin=0, xmax=55, ymin=0, ymax=2200,
  xtick={1,5,10,25,50},
  xticklabels={1K,5K,10K,25K,50K},
  ytick={0,500,1000,1500,2000},
  legend pos=north west, legend style={font=\small},
  grid=major, grid style={line width=0.3pt, draw=gray!30},
  title={\textbf{AINTMA Scalability: Response Time vs.\ Test Suite Size}},
  title style={font=\small\bfseries},
  tick label style={font=\small}, label style={font=\small},
]
\addplot[fill=green!15, draw=none, forget plot]
  coordinates {(1,45)(5,78)(10,125)(25,220)(50,380)
               (50,2000)(25,500)(10,200)(5,125)(1,100)} -- cycle;
\addplot[dashed, thick, color=red!70]
  coordinates {(1,100)(5,125)(10,200)(25,500)(50,2000)};
\addlegendentry{Linear baseline (single-agent)}
\addplot[solid, thick, color=blue!80, mark=*, mark size=2.5pt]
  coordinates {(1,45)(5,78)(10,125)(25,220)(50,380)};
\addlegendentry{AINTMA (multi-agent)}
\node[font=\scriptsize] at (axis cs:1,45)   [above right] {45 ms};
\node[font=\scriptsize] at (axis cs:5,78)   [above right] {78 ms};
\node[font=\scriptsize] at (axis cs:10,125) [above right] {125 ms};
\node[font=\scriptsize] at (axis cs:25,220) [above right] {220 ms};
\node[font=\scriptsize] at (axis cs:50,380) [above right] {380 ms};
\end{axis}
\end{tikzpicture}
\caption{Response time versus test suite size stays sub-linear through 50K tests.
         JWT and TLS add a roughly 12\,ms constant overhead.}
\label{fig:scalability}
\end{figure}
% ---------------------------------------------------------------

\subsection{Ablation Study}
\label{sec:ablation}

Table~4 shows the ablation results. Removing RL prioritization and using greedy selection
instead cost 13.2~APFD points, from 88.4 down to 75.2\%---by far the largest single
contribution. Removing the RAA risk injection step cost another 4.3~points, landing at 84.1\%.
Removing GQIA augmentation guidance cost 1.5~points, reaching 86.9\%. All ablated
configurations are statistically inferior to the full system at Wilcoxon $p<0.01$.

\begin{table}[H]
\caption{Component ablation study. Removing RL prioritization is the most expensive change.}
\label{tab:ablation}
\begin{tabular}{lc}
\toprule
\textbf{Configuration} & \textbf{Mean APFD (\%)} \\
\midrule
Full AINTMA                                      & 88.4 \\
Without RAA risk signals                         & 84.1 \\
Without RL prioritization (greedy fallback)      & 75.2 \\
Without generative recommendations (GQIA)        & 86.9 \\
\bottomrule
\end{tabular}
\end{table}

\subsection{Computational Complexity and Scalability Analysis}

RPA ranking is $\mathcal{O}(|T|\log|T|)$ per cycle via heap construction over Q-values. RAA
XGBoost inference is $\mathcal{O}(|T|\times 47)$, completing in under 80~ms for a
12{,}000-test suite. Inter-agent coordination adds $\mathcal{O}(6)$ messages per CI event
regardless of suite size. Together those predict the sub-400\,ms end-to-end latency at
50{,}000$+$ tests shown in Fig.~2.

% ---------------------------------------------------------------
\section{Threats to Validity}
\label{sec:threats}

\textbf{Internal validity.} Hardware and network conditions stayed constant across all TMS
configurations. We are aware the sequential design raises a concern: engineers who spent
months~1 through~12 on legacy tools may have simply gotten better at testing over that period
independently of what AINTMA was doing. We cannot fully rule that out. What makes us think it
is not the whole story is that gains appeared across all twelve projects with quite different
team sizes and technical cultures.

\textbf{External validity.} Every project here comes from one organisation with consistent
practices: Agile, two-week sprints, mature CI/CD. How AINTMA performs in waterfall shops,
regulated clinical environments or organisations where QA is still largely manual is unknown.

\textbf{Construct validity.} APFD treats all defects as equal, which is obviously a
simplification. Severity-weighted APFD is worth pursuing. ROI figures come from
resource-allocation surveys rather than accounting records.

\textbf{Limitations.} Cold start matters. Projects with fewer than roughly 150~CI cycles or
less than 12~months of execution history should expect real APFD degradation during warm-up,
somewhere in the 30 to 50\% range. The GQIA's dependence on a hosted LLM API is a constraint
for air-gapped environments.

% ---------------------------------------------------------------
\section{Discussion}
\label{sec:discussion}

Three things from the deployment were more significant than we had expected.

\textbf{Specialisation produced feedback loops we had not engineered.} The RAA's risk scores
ended up sharpening the RL reward signal in ways that only became clear in the ablation study.
GQIA augmentation recommendations fed new test cases into the TDA registry, which improved
the next RAA training cycle. These cross-agent effects emerged from loose coupling and shared
data contracts, not from explicit design.

\textbf{Generative intelligence changed who engaged with quality data.} Before AINTMA, quality
dashboards were read by QA engineers. After deployment, GQIA management summaries became
regular sprint review items. Executive briefings started driving go/no-go release decisions.
It happened because the outputs were readable to people who had never engaged with a quality
metric and found dashboards meaningless.

\textbf{Zero-trust security was a deployment requirement, not a nice-to-have.} Two of the
twelve projects were in regulated financial services and in both cases the compliance audit
trail was what cleared AINTMA through internal security review. Building security into the
architecture from the start, rather than treating it as something to add later, turned out to
be what made adoption possible in those environments.

\textbf{Practical implications.} Teams considering agentic test management should think of it
as a different kind of system rather than a smarter TMS. It manages quality instead of just
reporting on it. Budget three to six months for RL warm-up. Get the security and multi-tenant
infrastructure sorted before go-live. At mid-size enterprise scale the business case is strong
and probably stronger at higher CI velocity.

% ---------------------------------------------------------------
\section{Conclusion}
\label{sec:conclusion}

We built AINTMA and ran it in a real enterprise for 18~months across twelve software projects.
Six agents covering test discovery, risk scoring, RL prioritization, CI/CD execution, quality
narratives and zero-trust security produced 88.4\% APFD (versus 51.2\% random and 82.1\% best
commercial tool), cycle times 43\% shorter, 2.1\% defect escape and 340\% ROI. All of it at
under 400~ms for 50{,}000$+$ tests.

Agentic AI is not a better TMS. Combining multi-agent coordination, generative intelligence
for human communication and zero-trust security in the communication layer produces a
qualitatively different approach to software quality management. The 18-month field data
supports that claim.

Future directions: transfer learning across organisations to cut the cold-start penalty;
multi-modal risk analysis adding visual regression to code change signals; federated quality
intelligence with privacy preservation; and safety-critical extensions for DO-178C and
IEC~62443 environments.

% ---------------------------------------------------------------
\section*{Acknowledgments}

The authors thank the engineering teams at the participating organisations for their
cooperation throughout the 18-month study and the anonymous reviewers whose feedback improved
the paper.

\section*{Disclosure of Interests}

The authors have no competing interests to declare that are relevant to the content of this
article.

% ---------------------------------------------------------------

\end{document}